\documentclass[lettersize,journal]{IEEEtran}
\usepackage{amsmath,amsfonts}
\usepackage{algpseudocode}
\usepackage{algorithm}
\usepackage{array}
\usepackage[caption=false,font=normalsize,labelfont=sf,textfont=sf]{subfig}
\usepackage{textcomp}
\usepackage{url}
\usepackage{verbatim}
\usepackage{graphicx}
\usepackage{cite}
\usepackage{tabularx}
\usepackage{xcolor}
\usepackage{subcaption}
\usepackage{dblfloatfix}
\makeatletter
\renewenvironment{IEEEbiographynophoto}[1]{%
\if@IEEEbiographyTOCentrynotmade%
\setcounter{IEEEbiography}{-1}%
\refstepcounter{IEEEbiography}%
\addcontentsline{toc}{section}{Biographies}%
\global\@IEEEbiographyTOCentrynotmadefalse%
\fi%
\refstepcounter{IEEEbiography}%
\addcontentsline{toc}{subsection}{#1}%
\normalfont\@IEEEcompsoconly{\sffamily}\footnotesize\interlinepenalty500%
\vskip 1\baselineskip%
\parskip=0pt\par%
\noindent\textbf{#1\ }\@IEEEgobbleleadPARNLSP}{\relax\par\normalfont}
\makeatother

\begin{document}

\title{Explainable Reinforcement Learning for \\ Adaptive Traffic Signal Control}

\author{Dickens~Kwesiga,
        Nishu~Choudhary,
        Angshuman~Guin,
        and~Michael~Hunter% <-this % stops a space
\thanks{The authors are with the School of Civil and Environmental Engineering, Georgia Institute of Technology, Atlanta, GA 30332 USA (e-mail: dkwesiga@gatech.edu).}% <-this % stops a space
\thanks{Manuscript received June 30, 2026.}}

% The paper headers
\markboth{IEEE Transactions on Reliability}
{Kwesiga \MakeLowercase{\textit{et al.}}: Explainable Reinforcement Learning for Adaptive Traffic Signal Control}

%\IEEEpubid{0000--0000/00\$00.00~\copyright~2026 IEEE}
% Remember, if you use this you must call \IEEEpubidadjcol in the second
% column for its text to clear the IEEEpubid mark.

\maketitle

\begin{abstract}
Model-free Reinforcement Learning (RL) has emerged as a powerful paradigm for adaptive traffic signal control. This is partly because state-of-the-art RL agents can directly interact with traffic simulation environments to learn highly complex, non-linear traffic control policies without relying on complex predictive models. However, in safety-critical infrastructure like traffic control, the opaque, black-box nature of deep RL models poses challenges for transportation agency acceptance, regulatory compliance, operational trust, troubleshooting, and fine-tuning by traffic engineers. To bridge this gap between high-performance optimization and human-comprehensible interpretability, this effort introduces a novel, explainable entity-centric RL framework for safe and transparent traffic signal control. Rather than processing traffic states through monolithic, flat vectors, the proposed architecture disaggregates real-time intersection observations into distinct, high-dimensional lane entities (acting as Queries) and phase temporal configurations (acting as Keys and Values) to inherently preserve the structural topology and geometric configurations of the intersection. Relational dependencies and inter-lane conflicts are dynamically extracted via a dual-stage attention network featuring sequential multi-head cross-attention and self-attention blocks. This design yields a real-time affinity matrix that quantifies the direct influence of signal phases on specific approach volumes and queues, providing full visual and analytical interpretability. To ensure strict operational reliability, a deterministic action-masking interface is integrated directly into the Proximal Policy Optimization (PPO) pipeline, explicitly blocking invalid phase transitions to guarantee absolute compliance with established signal timing norms and safety constraints. Evaluated in a microscopic simulation environment under diverse traffic demands, the framework matches or outperforms state-of-the-art baselines in delay minimization. More importantly, the emergent attention weights align precisely with established traffic engineering principles, tracking queue clearance and coordinated phase transitions, offering an auditable, trust-enabling, and deployable architecture for next-generation adaptive traffic control systems. 
\end{abstract}


\section{Introduction}
\IEEEPARstart{M}{}odel-free Reinforcement Learning (RL) has emerged as a powerful paradigm for adaptive traffic signal control. By interacting directly with microscopic traffic simulation environments, state-of-the-art RL agents can learn highly complex, non-linear optimization policies without relying on complex predictive models. These frameworks map real-time traffic observations directly to optimal signaling decisions, consistently demonstrating superior performance over conventional actuated and fixed-time controllers in simulated environments.

Despite the demonstrated successes of RL-based traffic signal controllers in literature, the opaque, black-box nature of deep RL models poses challenges for transportation agency acceptance, operational trust, troubleshooting, and fine-tuning by traffic engineers. Previous deep RL models produce control actions without exposing the underlying causal logic or structural justifications behind their decisions. In safety-critical infrastructure such as traffic control, the absence of explainability introduces significant barriers to deployment, including challenges in regulatory compliance and operational trust. Additionally, black-box models offer no structural diagnostics to distinguish between anomalous sensor inputs, hardware failures and poor policy inference limiting the ability of traffic engineering teams to easily troubleshoot. Furthermore, RL algorithms with obscured internal logic make it impossible for engineers to fine-tune or inject domain knowledge into the control process. There is a clear need for an interpretable, structurally organized RL framework that bridges the gap between high-performance deep optimization and human-comprehensible explainability. 

This paper introduces an explainable, entity-centric RL framework designed specifically for safe and transparent traffic signal control. By shifting from flat vector spaces to a decoupled, high-dimensional entity representation, the proposed architecture reveals the underlying dependencies between traffic movements and signal phases that drive control decisions.

The primary contributions of this effort are as follows:
\begin{enumerate}
\item{Topology-Preserving Entity Embedding: The study proposes an architectural decomposition that isolates raw intersection metrics into individual lane entities (Queries) and phase status elements (Keys/Values), mapping them into a shared high-dimensional latent space to inherently preserve the spatial configuration and spatial topology of the intersection.}
\item{Hierarchical Relational Attention Mechanism: A dual-stage attention architecture combining multi-head cross-attention and multi-head self-attention is introduced to model interactions between lanes and signal phases, as well as inter-lane dependencies. This structure provides fully explainable matrices that quantify the direct influence of each signal phase on individual approach lanes, illuminating the model's internal decision-making logic in real time.}
\item{Constrained Action Masking Interface: To bridge the gap between stochastic policy exploration and deterministic field safety, the formulated architecture integrates an analytical action-masking layer directly into the RL policy pipeline. This interface mathematically blocks invalid phase transitions, guaranteeing absolute compliance with established signal timing constraints.}
\item{Empirical Validation and Explainability Assessment: Utilizing the Simulation of Urban MObility (SUMO) environment, the proposed approach demonstrates improved performance in delay reduction compared to baseline methods, while providing interpretable attention visualizations that align with observed traffic phenomena such as queue formation, phase transitions, and spillback mitigation.}
\end{enumerate}

\section{RELATED STUDIES}
\subsection{RL for Traffic Signal Control}
A substantial body of recent work has focused on developing RL-based adaptive signal control systems \cite{1,2,3,4,5,6,7,8,9,10,11,12,13,14,15,16,17,18}. The appeal of RL-based signal control lies in the model free nature of state-of-the-art RL algorithms. Unlike the current field deployed adaptive signal control systems, model free RL-based adaptive signal control systems do not rely on state predictive models, making them more computationally efficient for real-time implementation. In simulated environments, RL-based has shown superior performance to the state-of-practice fixed time and actuated signal control systems.

Early research efforts largely focused on isolated intersection control, formulating single RL agents trained in microscopic traffic simulation to optimize signal timings at individual intersections \cite{13,19,20,21,22,23,24,25,26,27}. Several studies propose algorithms based on the deep Q-network (DQN) framework and its extensions \cite{8,9,13,21,24,25,27,28,29,30} while others propose algorithms based on policy-based methods including actor-critic and its variations such as advantage actor-critic (A2C), deep deterministic policy gradient (DDPG), and proximal policy optimization (PPO) \cite{19,22,31,32}.

More recent efforts have focused on extending single agent RL-based signal control to multi agent reinforcement learning (MARL)-based signal control with a single agent controlling each intersection and cooperating with the adjacent intersection agents to generate a coordinated signal timing plan \cite{1,2,5,6,7,10,11,14,15,16,33,34,35,36}. Several efforts adopt the centralized training and decentralized execution (CTDE) paradigm of MARL which allows implicit communication between the agents during training. Some of these efforts enhance the implicit communication inherent in CTDE with explicit communication modules to exchange observation and action histories across agents \cite{6,11,18}. Some studies especially for network level signal control have formulated graph based RL \cite{1,4,7,10}. Graph based RL allows information sharing between neighboring agents through graph neural networks.

\subsection{Explainable RL for Traffic Signal Control}
Recent work has begun to address the interpretability limitations of RL-based traffic signal control. Luo, et al. \cite{37} proposed an interpretable influence mechanism based on efficient hinging hyperplanes neural networks, which leverages ANOVA decomposition to quantify the contribution of individual traffic features and their interactions. This approach enables explicit estimation of feature importance and provides an interpretation of spatiotemporal dependencies among intersections. However, while the method offers strong analytical interpretability, it lacks the intuitive and visual explainability and do not explicitly capture structured relationships between traffic movements and signal phases.

Hu, et al. \cite{38} proposed an explainable RL framework that integrates an attention-based deep Q-network to model vehicle-level interactions and prioritize critical vehicles in decision-making. The study directly encodes individual vehicle attributes, enabling more fine-grained control and improved transparency through attention visualization and counterfactual analysis. Experimental results demonstrated significant improvements in travel time and queue reduction compared to conventional and RL-based baselines, while also providing interpretable insights into the learned control policies. Despite these advances, the reliance on vehicle-level representations introduces challenges in scalability and compatibility with commonly available traffic sensing infrastructure, which is typically aggregated at the lane or movement level. Furthermore, the use of value-based methods limits flexibility in handling complex action spaces and temporal dependencies.
\subsection{Attention Mechanisms for Explainability}
Originally introduced by Bahdanau \emph{et al.}~\cite{39} to overcome the information bottleneck in machine translation, the attention mechanism was designed to dynamically map dependencies between input and output sequences. Subsequent architectural integrations specifically combining scaled dot-product attention with residual connections~\cite{40}, positional encodings, and layer normalization~\cite{41} culminated in the Transformer architecture developed by Vaswani \emph{et al.}~\cite{42}. Beyond its foundational dominance in natural language processing (NLP), the Transformer framework has become ubiquitous across diverse machine learning domains, including computer vision~\cite{43}, structural biology~\cite{44}, and molecular design~\cite{45}.

In addition to driving state-of-the-art predictive performance, attention weights are increasingly leveraged as an intrinsic layer of model interpretability. Because these weights provide a quantifiable distribution of a model's focus across spatial or temporal inputs, they offer a window into the otherwise opaque decision-making processes of deep architectures. This explainability aspect has been utilized across several fields, including medical diagnostic coding using reverse-time dual attention mechanisms~\cite{46, 47}, and image captioning, where spatial attention heatmaps are superimposed to visualize the exact pixel regions driving textual outputs~\cite{48}. 

Crucially, this interpretive capacity has extended into explainable RL. In safety-critical systems, attention-augmented agents can map internal policies directly back to observable environmental states~\cite{49}.

\section{Problem Formulation}
This section presents the formulation of the proposed attention-based RL framework for traffic signal control. The problem is formulated as a semi-Markov decision process (SMDP) to account for the event-driven nature of signal control, where actions are executed over variable time intervals. To enable structured decision-making and interpretability, the traffic environment is represented using an entity-based state formulation that explicitly captures lane-level traffic conditions and signal phase information. These entities are processed through attention mechanisms to model interactions between traffic demand and control actions.

\begin{figure*}[!t]
    \centering
    \includegraphics[width=\linewidth]{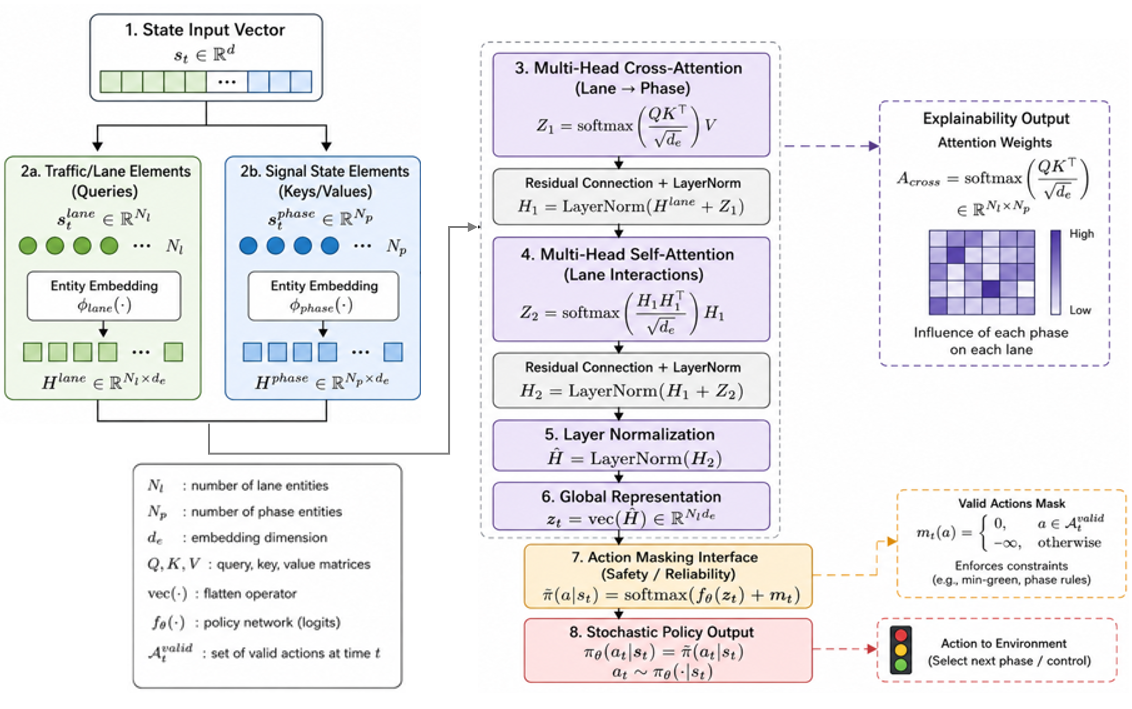}
    \caption{Overview of the proposed attention-based RL architecture for traffic signal control.}
    \label{fig:fig_1}
\end{figure*}

A high-level overview of the proposed formulation and architecture is illustrated in Fig.~\ref{fig:fig_1}. The key components of the formulation, including state representation, attention mechanism, action space, reward definition, and policy structure, are described in the following subsections.

\subsection{State Space Definition}
To effectively capture the real-time intersection dynamics and structural 
topologies, the state space is modeled using an entity-centric representation 
rather than a flat, monolithic vector. At each decision time-step $t$, the global state vector $s_t\in\mathbb{R}^d$, aggregates traffic state $s_t^{lane}$ and signal state $s_t^{phase}$:
\begin{equation}
    s_t = \left[ s_t^{\text{lane}} \parallel s_t^{\text{phase}} \right]
    \label{eq:state_aggregation}
\end{equation}
where $\parallel$ denotes the concatenation operator and $d=\ N_l+N_p$ represents the total dimensionality composed of $N_l$ lane entities and $N_p$ phase entities.
\subsubsection{State Input Space Decomposition}
The state vector is cleanly partitioned into two distinct functional categories:
\begin{itemize}
    \item \textbf{Traffic/Lane Elements (Queries):} The vector $s_t^{\text{lane}} = \left[ v_t^1, v_t^2, \dots, v_t^{N_l} \right] \in \mathbb{R}^{N_l}$ captures the intersection traffic demand. Each element $v_t^i$ represents the vehicle count for lane entity $i$. Within the relational learning mechanism, these elements serve as Queries ($Q$), prompting the policy to evaluate per-lane demand distribution against signal phasing and green time allocation.
    \item \textbf{Signal State Elements (Keys/Values):} The vector $s_t^{\text{phase}} = \left[ g_t^1, g_t^2, \dots, g_t^{N_p} \right] \in \mathbb{R}^{N_p}$ represents the temporal state of the traffic signal controller. Each element $g_t^i$ tracks the green duration of phase $i$. These elements serve as Keys ($K$) and Values ($V$), providing the contextual grid that bounds the traffic/vehicle metrics.
\end{itemize}
\subsubsection{Disaggregated High-Dimensional Entity Embedding}
Because the global state vector $s_t$ aggregates raw features across disparate traffic components (per lane vehicle counts vs. phase durations), feeding a flat vector directly into standard neural layers risks obfuscating the spatial and functional relationships unique to each lane and phase. To preserve structural topology, $s_t$ is decomposed into separate, distinct entity representations and map them into a shared high-dimensional space $\mathbb{R}^{d_e}$, where $d_e$ is the embedding dimension.

Let $X_{l,i}\in\mathbb{R}^1$ represent the raw feature scalar (vehicle count $v_t^i)$ of lane i\ and let $X_{p,j}\in\mathbb{R}^1$ represent the raw feature scalar (green duration $g_t^j$) of phase j. Using parameterized, entity-specific mapping layers (implemented via specialized dense layers with a Rectified Linear Unit activation), each individual entity is projected into the latent space:
\begin{equation}
    \mathbf{h}_i^{\text{lane}} = \text{ReLU}\left(\mathbf{W}_i^{\text{lane}} X_{l,i} + b_i^{\text{lane}}\right) \in \mathbb{R}^{d_e}
    \label{eq:lane_project}
\end{equation}
\begin{equation}
    \mathbf{h}_j^{\text{phase}} = \text{ReLU}\left(\mathbf{W}_j^{\text{phase}} X_{p,j} + b_j^{\text{phase}}\right) \in \mathbb{R}^{d_e}
    \label{eq:phase_project}
\end{equation}
where $\mathbf{W}_i^{\text{lane}}$, $\mathbf{W}_j^{\text{phase}}$ represent the trainable projection matrices, and $b_i^{\text{lane}}$, $b_j^{\text{phase}}$ denote their respective bias vectors.

Following individual projections, the high-dimensional representations are stacked along a new entity dimension to produce the final structured matrices passed to the attention heads:
\begin{align}
    H^{\text{lane}} &= \begin{bmatrix}
        \text{---} & h_1^{\text{lane}} & \text{---} \\
        & \vdots & \\
        \text{---} & h_{N_l}^{\text{lane}} & \text{---}
    \end{bmatrix} \in \mathbb{R}^{N_l \times d_e} \label{eq:matrix_lane} \\[1ex]
    H^{\text{phase}} &= \begin{bmatrix}
        \text{---} & h_1^{\text{phase}} & \text{---} \\
        & \vdots & \\
        \text{---} & h_{N_p}^{\text{phase}} & \text{---}
    \end{bmatrix} \in \mathbb{R}^{N_p \times d_e} \label{eq:matrix_phase}
\end{align}
By decoupling the entities in this manner, $H^{lane}$ operates strictly as an array of structured Queries, while $H^{phase}$ operates as Keys and Values. This guarantees that the subsequent cross-attention layer computes an explicit affinity matrix capturing the exact relational weight between every single approach lane and every traffic phase configuration.
\subsection{Relational Representational Learning Via Hierarchical Attention}
To extract structural dependencies between lanes and phases without assuming a fixed geometric intersection layout, the high-dimensional entity matrices $H^{lane}$ and $H^{phase}$ are passed through a hierarchical attention network. This pipeline explicitly models directed lane-to-phase dependencies followed by localized lane-to-lane interactions.
\subsubsection{Multi-Head Cross-Attention (Lanes - Phase)}
The first stage establishes contextual relationships between the traffic demand on individual lanes and the operating configurations of the traffic phases. The lane embedding matrix $H^{lane}$ is projected to form the Queries (Q) while the phase embedding matrix $H^{phase}$ is projected to form the Keys (K) and Values (V):
\begin{equation}
    Q = H^{\text{lane}} W_Q, \quad K = H^{\text{phase}} W_K, \quad V = H^{\text{phase}} W_V
    \label{eq:qkv_projections}
\end{equation}
where $W_Q$, $W_K$, $W_V\ \in\  \mathbb{R}^{{d_eXd}_e}$ are trainable projection matrices. The cross-attention output matrix $Z_1$ is computed using a scaled dot-product formulation:
\begin{equation}
    Z_1 = \text{Attention}(Q, K, V) = \text{softmax}\left( \frac{QK^T}{\sqrt{d_e}} \right) V
    \label{eq:attention_mechanism}
\end{equation}

The intermediate attention weight matrix produced by this block will be used to generate attention maps for explain ability. The cross attention $A_{cross}$ acts as a verifiable record of agent’s reasoning. It mathematically represents the exact relational importance assigned to every individual lane relative to each available green signal combination.
\begin{equation}
    A_{\text{cross}} = \text{softmax}\left( \frac{QK^T}{\sqrt{d_e}} \right) \in \mathbb{R}^{N_l \times N_p}
    \label{eq:cross_attention_matrix}
\end{equation}

To stabilize training gradients and mitigate information loss, a residual connection followed by layer normalization is applied to generate the first intermediate latent state $H_1$:
\begin{equation}
    H_1 = \text{LayerNorm}\left( H^{\text{lane}} + Z_1 \right) \in \mathbb{R}^{N_l \times d_e}
    \label{eq:layernorm_residual}
\end{equation}
\subsubsection{Multi-Head Self-Attention (Lane - Lane interactions)}
While cross-attention captures lane-to-phase dynamics, lanes approaching an intersection also experience inter-lane dependencies (e.g., merging conflicts or queue spillbacks). To capture these localized conflicts, $H_1$ is passed through a multi-head self-attention layer where queries, keys, and values are derived entirely from the updated lane representations:
\begin{equation}
    Z_2 = \text{softmax}\left( \frac{H_1 H_1^T}{\sqrt{d_e}} \right) H_1
    \label{eq:self_attention_z2}
\end{equation}
A second residual connection and layer normalization step are executed to form the interactive lane representation matrix $H_2$
\begin{equation}
    H_2 = \text{LayerNorm}\left( H_1 + Z_2 \right) \in \mathbb{R}^{N_l \times d_e}
    \label{eq:layernorm_residual_2}
\end{equation}
\subsubsection{Global Representation and Layer Normalization}
Prior to generating the final policy distribution, the latent matrix undergoes an additional layer normalization step to ensure numerical stability against extreme variations in traffic patterns:
\begin{equation}
    \widehat{H} = \text{LayerNorm}\left( H_2 \right) \in \mathbb{R}^{N_l \times d_e}
    \label{eq:final_layernorm}
\end{equation}

Finally, a vectorization operator $(\odot)$ flattens the structured entity matrix across its lane and latent dimensions. This generates a compact, global feature vector $Z_t$ capturing the comprehensive structural and traffic state of the intersection:
\begin{equation}
    Z_t = \text{vec}\left( \widehat{H} \right) \in \mathbb{R}^{N_l \cdot d_e}
    \label{eq:vectorization}
\end{equation}

This unified representation $Z_t$ encapsulates both traffic demand distributions and signal timing context, serving as the direct input for the downstream action-masking and stochastic policy layers.
\subsection{Safety Constrained Action Space}
\begin{figure}[!h]
    \centering
    \includegraphics[width=0.5\linewidth]{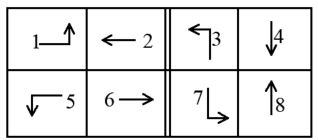}
    \caption{Ring-barrier diagram}
    \label{fig:fig_2}
\end{figure}

In this formulation, action is defined based on the NEMA dual ring-barrier diagram shown in Fig.~\ref{fig:fig_2} A discrete set of eight actions (0-7) is defined with each action corresponding to a pair of compatible phases, one from each ring. Actions 0 to 7 correspond to phase combinations (1,5), (1,6), (2,5), (2,6), (3,7), (3,8), (4,7), (4,8).
Because of the constraints of minimum green, yellow and red clearance intervals and double barrier crossing requirements, not all the eight actions are available for selection at every decision point. At every decision step t, the algorithm evaluates the valid action set $A_t^{valid}\subset A$ as a function of active phase, committed phase and elapsed phase duration.

The raw policy network processes the geometric feature vector $Z_t$ to output policy logits $f_\theta(Z_t) \in \mathbb{R}^{N_{\text{actions}}}$. The outputs are filtered by a deterministic safety mask $m_t \in \mathbb{R}^{|A|}$. This mask encodes structural traffic rules including minimum green, yellow, and red clearances, and NEMA dual-ring barrier constraints at runtime:
\begin{equation}
    m_t(a) = \begin{cases}
        0, & \text{if } a \in A_t^{\text{valid}} \\
        -\infty, & \text{if } a \notin A_t^{\text{valid}}
    \end{cases}
    \label{eq:action_mask}
\end{equation}
To explicitly eliminate the probability of selecting an invalid and unsafe phase change, the mask vector $m_t$ is added directly to the raw network logits. This penalizes invalid choices prior to normal distribution scaling:
\begin{equation}
\begin{split}
    \tilde{\pi}(a \mid s_t) &= \text{softmax}\left( f_{\theta}(Z_t) + m_t \right) \\
    &= \frac{\exp\left( l_t(a) + m_t(a) \right)}{\sum_{j \in A} \exp\left( l_t(a_j) + m_t(j) \right)}
\end{split}
\label{eq:policy_distribution}
\end{equation}
By setting the logits of invalid movements to $-\infty$, their resulting exponentiated values evaluate cleanly to zero, ensuring that the filtered distribution $\widetilde{\pi}\left(a\middle| s_t\right)$ assigns zero selection probability to any phase that violates ring-barrier safety constraints. The policy probability $\pi_\theta(a_t|s_t)$ maps directly to the valid action space:
\begin{equation}
    \pi_{\theta}(a_t \mid s_t) = \tilde{\pi}(a_t \mid s_t)
    \label{eq:policy_mapping}
\end{equation}
The actual discrete action $a_t$ selected for implementation is sampled directly from this distribution:
\begin{equation}
    a_t \sim \pi_{\theta}(\cdot \mid s_t)
    \label{eq:action_sampling}
\end{equation}
This action dictates the next phase configuration or phase extension step to apply over the subsequent control interval, combining high-dimensional representation learning with strict operational guardrails.
\subsection{Reward Definition}
To optimize the efficiency of the intersection, the objective of the RL is to minimize cumulative vehicular delay. The reward function $r_t$ at any decision step t is formulated as the negative sum of normalized delays experienced by all active vehicles across the monitored approach lanes:
\begin{equation}
    r_t = -\sum_{l \in N_l} \sum_{v \in V_l(t)} \frac{d_v(t)}{D}
    \label{eq:traffic_reward}
\end{equation}
where $d_v$ is delay for individual vehicle v for all vehicles $V_l$ on lane l at a time t.\ D is the normalization scaling factor. The individual vehicle delay $d_v(t)$ is calculated analytically as the difference between the total time spent by vehicle $v$ in the network and its theoretical free-flow travel time
\subsection{Proximal Policy Optimization Formulation}
The relational feature vector $Z_t$ and filtered distribution $\pi_\theta\left(a_t\middle| s_t\right)$ are optimized using Proximal Policy Optimization (PPO), a policy gradient variant formulated by \cite{39} that stabilizes training by constraining policy updates via an analytical clipping mechanism. The agent is structured as an Actor-Critic architecture, updating a parameterized policy network $\pi_\theta$ and a state-value function $V_\vartheta$.
\subsection{Generalized Advantage Estimation (GAE)}
To minimize variance during trajectory evaluation, the generalized advantage estimator $\tilde{A}_t$ is computed via temporal difference errors $\delta_t^V$ across a discounted reverse-accumulation buffer:
\begin{equation}
    \delta_t^V = r_t + \gamma V_\vartheta(s_{t+1})(1 - d_t) - V_\vartheta(s_t)
    \label{eq:td_error}
\end{equation}

\begin{equation}
    \tilde{A}_t = \sum_{k=0}^{T-t-1} (\gamma \lambda)^k \delta_{t+k}^V, \quad \tilde{A}_t^{\text{norm}} = \frac{\tilde{A}_t - \mu_{\tilde{A}}}{\sigma_{\tilde{A}} + \epsilon}
    \label{eq:gae_computation}
\end{equation}
where $r_t$ is the reward, $\gamma$ is the discount factor, $\lambda$ is the GAE trace decay hyperparameter, and $d_t$ is the terminal indicator.

\subsection{Clipped Surrogate Policy Objective}
To prevent extreme policy updates that disrupt the delicate attention mappings, PPO enforces a probability ratio constraint. Let $\rho_t(\theta)$ define the relative probability tracking parameter between the updated policy $\pi_\theta$ and the old policy $\pi_{\theta_{\text{old}}}$ used to collect transitions:
\begin{equation}
    \rho_t(\theta) = \frac{\pi_\theta(a_t \mid s_t)}{\pi_{\theta_{\text{old}}}(a_t \mid s_t)}
    \label{eq:probability_ratio}
\end{equation}

The surrogate policy objective function minimizes a conservative lower bound, clipping the ratio if it moves outside an acceptable trust region governed by hyperparameter $\epsilon_{\text{clip}}$:
\begin{equation}
\begin{split}
    \mathcal{L}^{\text{CLIP}}(\theta) = \hat{\mathbb{E}}_t \Big[ \min \Big( &\rho_t(\theta)\hat{A}_t^{\text{norm}}, \\
    &\text{clip}\left(\rho_t(\theta), 1 - \epsilon_{\text{clip}}, 1 + \epsilon_{\text{clip}}\right)\hat{A}_t^{\text{norm}} \Big) \Big]
\end{split}
\label{eq:ppo_clip_split}
\end{equation}

To maintain policy diversity and prevent premature convergence to a singular signal sequence, an entropy regularization term $\mathcal{H}(\pi_\theta(\cdot \mid s_t))$ is introduced with an allocation coefficient $c_2$:
\begin{equation}
    \mathcal{L}^{\text{actor}}(\theta) = \mathcal{L}^{\text{CLIP}}(\theta) + c_2 \hat{\mathbb{E}}_t \left[ \mathcal{H}\left(\pi_\theta(\cdot \mid s_t)\right) \right]
    \label{eq:actor_loss}
\end{equation}
\section{Model training}
\subsection{Network Model}
\begin{figure}[!h]
    \centering
    \includegraphics[width=1\linewidth]{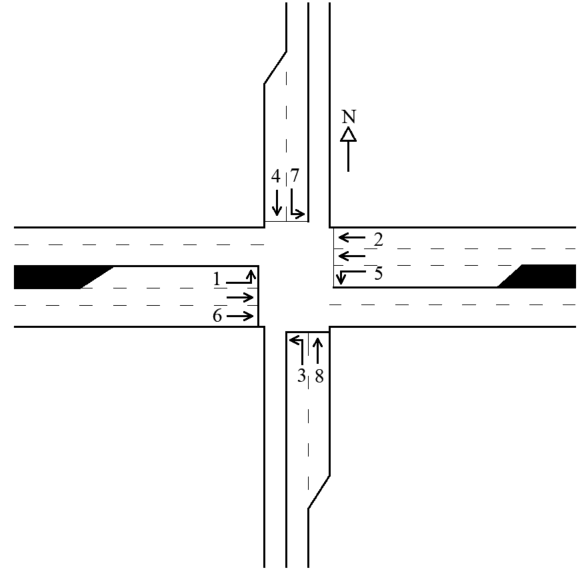}
    \caption{Schematic representation of the simulated intersection geometry}
    \label{fig:fig_3}
\end{figure}

The model is trained for a single intersection network illustrated in Fig.~\ref{fig:fig_3}. The network consists of a four-leg signalized intersection with a major street (east–west) and a minor street (north–south). The main street is configured with two through lanes in each direction and a short exclusive left-turn pocket, while the side street consists of one through lane in each direction with a short left-turn pocket. 

The model is trained on three distinct demand patterns (A, B, C) as shown in Table~\ref{tab:od_patterns}. These volume patterns are selected to be significantly different from one another, as quantified using the structural similarity index formulated by Wang, et al. \cite{51}. Exposing the model to diverse traffic demand conditions is expected to improve its robustness and ability to generalize to unseen volumes.

\begin{table}[!h]
\caption{O-D Patterns Used to Train the Model}
\label{tab:od_patterns}
\centering
\renewcommand{\arraystretch}{1.2}
\setlength{\tabcolsep}{3pt} % Squeezes horizontal spacing between columns to prevent bleed
\footnotesize
\begin{tabularx}{\columnwidth}{@{}c|XXXXXXXX@{}}
\hline
\textbf{O-D Pattern} & \textbf{EBL} & \textbf{EBT} & \textbf{WBL} & \textbf{WBT} & \textbf{NBL} & \textbf{NBT} & \textbf{SBL} & \textbf{SBT} \\
\hline
A & 171 & 1440 & 171 & 1440 & 157 & 270 & 257 & 270 \\
B & 250 & 1440 & 150 & 900  & 50  & 400 & 257 & 270 \\
C & 150 & 900  & 250 & 1440 & 257 & 270 & 50  & 400 \\
\hline
\end{tabularx}
\end{table}
\subsection{Model Interaction with Simulation Environment and Updating Network parameters }
\subsubsection{Distributed Simulation Environments}
Simulation of urban MObility (SUMO) is selected as the simulation environment \cite{52}. The model is trained in distributed simulation environments. As detailed in Algorithm~\ref{alg:multi_env_ppo}, training occurs over E episodes across N independent SUMO simulation environments running in parallel. At the beginning of an episode, each environment $i$ initializes its local simulation instance and constructs an empty trajectory storage buffer $B_i$. Over the simulation horizon $T$, step-by-step state tracking and control execution are offloaded to the event-driven routine described in Algorithm~\ref{alg:event_rl}.

Once the trajectories are collected, the sample buffers are combined into an aggregated dataset $D=\ \bigcup_{i=1}^{N}B_i$. GAE is then performed to calculate the target returns $R_t$ and standardized advantages ${\widetilde{A}}_t$ ensuring low-variance gradient trajectories during optimization updates.

\begin{algorithm}[!h]
\caption{Multi-Environment PPO Training}
\label{alg:multi_env_ppo}
\begin{algorithmic}[1]
\Require Number of environments $N$, episodes $E$, horizon $T$
\State Initialize policy $\pi_\theta$ and value network $V_\phi$

\For{$e=1,\ldots,E$}

\Statex \textcolor{blue}{\textbf{// Parallel Rollout Collection}}

\ForAll{environments $i=1,\ldots,N$ \textbf{(in parallel)}}
    \State Reset environment and initialize buffer $\mathcal{B}_i$
    \State Observe initial state $s_0$
    \For{$t=1,\ldots,T$}
        \State Advance simulation by one step
        \State Execute Algorithm~\ref{alg:event_rl}
        \State Store $(s_t,a_t,r_t,V_\phi(s_t),\log\pi_\theta(a_t|s_t))$ in $\mathcal{B}_i$
    \EndFor
    \State Store terminal transition
\EndFor

\Statex \textcolor{blue}{\textbf{// Batch Aggregation}}

\State $\mathcal{D}\leftarrow\bigcup_{i=1}^{N}\mathcal{B}_i$

\Statex \textcolor{blue}{\textbf{// Advantage Estimation}}

\State Compute returns $R_t$
\State Compute generalized advantages $\hat A_t$
\State Normalize $\hat A_t$

\Statex \textcolor{blue}{\textbf{// Policy Update}}

\For{$k=1,\ldots,K_\pi$}
    \State Compute probability ratio
    \[
    r_t(\theta)=
    \frac{\pi_\theta(a_t|s_t)}
         {\pi_{\theta_{\rm old}}(a_t|s_t)}
    \]
    \State Maximize clipped PPO objective with entropy bonus
    \State Update policy parameters $\theta$
\EndFor

\Statex \textcolor{blue}{\textbf{// Value Function Update}}

\For{$k=1,\ldots,K_V$}
    \State Minimize
    $
    L_V=\mathbb{E}\!\left[(V_\phi(s_t)-R_t)^2\right]
    $
    \State Update value parameters $\phi$
\EndFor

\Statex \textcolor{blue}{\textbf{// Checkpoint}}

\State Save policy $\pi_\theta$ and value network $V_\phi$

\EndFor

\Ensure Trained policy $\pi_\theta$ and value network $V_\phi$
\end{algorithmic}
\end{algorithm}

\begin{algorithm}[!h]
\caption{Event-Driven RL-Based Signal Control}
\label{alg:event_rl}
\begin{algorithmic}[1]

\Require Current state $s_t$, policy $\pi_\theta$
\State Initialize decision time $t_{\mathrm{decision}}$ and update time $t_{\mathrm{update}}$

\While{simulation is running}

    \State Advance simulation by one simulation step
    \State Update traffic state and vehicle tracking

    \Statex \textcolor{blue}{\textbf{// Decision Event}}

    \If{$t=t_{\mathrm{decision}}$}
        \State Determine valid action set $\mathcal{A}_t$
        \State Sample action $a_t\sim\pi_\theta(a|s_t,\mathcal{A}_t)$
        \State Estimate state value $V_\phi(s_t)$
        \State Compute action duration $\Delta t=f(a_t,\text{signal state})$
        \State $t_{\mathrm{update}}\gets t+\Delta t$
    \EndIf

    \Statex \textcolor{blue}{\textbf{// Action Execution}}

    \State Apply signal control corresponding to $a_t$
    \State Maintain the selected phase during $[t,t_{\mathrm{update}})$
    \State Enforce minimum green, yellow, and all-red intervals

    \Statex \textcolor{blue}{\textbf{// Transition Event}}

    \If{$t=t_{\mathrm{update}}$}
        \State Observe next state $s_{t+1}$
        \State Compute reward $r_t$
        \State Store transition
        \State $(s_t,a_t,r_t,V_\phi(s_t),\log\pi_\theta(a_t|s_t),s_{t+1})$
        \State $t_{\mathrm{decision}}\gets t+\epsilon$
        \State $s_t\gets s_{t+1}$
    \EndIf

\EndWhile

\end{algorithmic}
\end{algorithm}
\subsubsection{Event-Driven RL Control Loop}
The real-time interaction between the agent and the simulation is governed by the SMDP in Algorithm~\ref{alg:event_rl}. Decisions are triggered dynamically at $t = t_{\text{decision}}$. The environment uses the action masking layer to isolate a valid action set $A_t$ based on the current signal state. A safe action $a_t$ is sampled from the masked distribution $\pi_\theta(a_t \mid s_t, A_t)$, and its continuous operational duration $\Delta t = f(a_t, \text{signal state})$ is computed analytically.

The selected action is implemented in the environment over the variable window $(t, t_{\text{update}} = t + \Delta t)$ enforcing required intermediate intervals (such as yellow clearance and all-red safety flags). At $t_{\text{update}}$, a transition event updates the target parameters, records the trajectory tuple $(s_t, a_t, r_t, V_\vartheta(s_t), \log \pi_\theta(a_t \mid s_t))$ into buffer $B_i$, and increments the next look-ahead decision time step.

\subsubsection{Policy and Value Network Optimization}
Following data aggregation at the end of each episode, the policy parameters $\theta$ are updated over $K_\pi$ epochs. At each epoch, the clipped surrogate loss $\mathcal{L}^{\text{CLIP}}$ balances policy stability against historic target behavior while an entropy regularization term preserves exploration diversity. Concurrently, the critic value parameters $\vartheta$ are adjusted over $K_v$ epochs to minimize baseline prediction error. After structural optimization is complete, a network parameter checkpoint is saved, updating the global models across the worker environments.

\subsubsection{Hyper parameter selection}
Preliminary training was performed to select architectural and RL hyper parameters. The final selected hyper parameters are shown in Table 2.
\begin{table}[!h]
\caption{Training and Architectural Hyperparameters}
\label{tab:hyperparameters}
\centering
\renewcommand{\arraystretch}{1.2}
\footnotesize
\begin{tabularx}{\columnwidth}{l|>{\raggedright\arraybackslash}X|c|c}
\hline
\textbf{Parameter Category} & \textbf{Hyperparameter Description} & \textbf{Symbol} & \textbf{Value} \\
\hline
Entity Embedding & High-dimensional embedding dimensions & $d_e$ & 32 \\
Layer            & Embedding layer activation & $\sigma(\cdot)$ & ReLU \\
\hline
Attention Blocks & Number of cross-attention heads & $H_c$ & 4 \\
                 & Number of self-attention heads & $H_s$ & 4 \\
\hline
Policy/Critic    & Fully connected hidden layers & -- & 256, 128 \\
Networks         & Network activation function & -- & tanh \\
\hline
PPO Core         & Discount factor & $\gamma$ & 0.99 \\
Parameters       & Trace decay parameter (GAE) & $\lambda$ & 0.95 \\
                 & PPO clipping ratio & $\epsilon_{\mathrm{clip}}$ & 0.2 \\
                 & Entropy coefficient & $c_2$ & 0.01 \\
                 & Target KL divergence & $\mathrm{KL}_{\mathrm{target}}$ & 0.01 \\
\hline
Optimization     & Actor learning rate & $\alpha_\theta$ & $1\times10^{-4}$ \\
Steps            & Critic learning rate & $\alpha_\phi$ & $1\times10^{-4}$ \\
                 & Actor/Critic optimization epochs & $K_\pi, K_V$ & 80 \\
                 & Gradient clipping threshold & $g_{\mathrm{clip}}$ & 0.5 \\
\hline
Simulation       & Parallel environments & $N$ & 3 \\
Setup            & Episode horizon & $T$ & 3600 s \\
\hline
\end{tabularx}
\end{table}
\section{Results and Discussion}
\subsection{Model Training}
\begin{figure}[!h]
    \centering
    \includegraphics[width=1\linewidth]{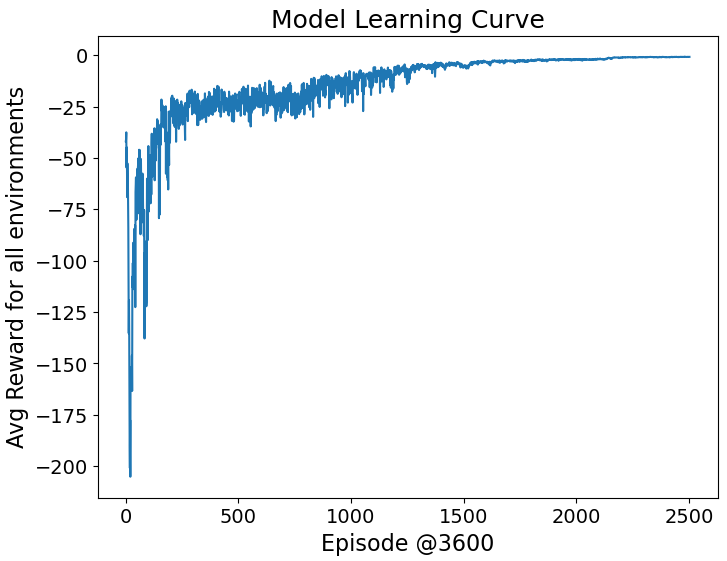}
    \caption{Model Learning Curve}
    \label{fig:fig_4}
\end{figure}

Fig.~\ref{fig:fig_4} illustrates the training performance of the proposed PPO agent, showing the evolution of the average episodic reward aggregated across all three distributed environments. The reward is defined as the negative normalized delay, computed from vehicle travel times relative to estimated free-flow conditions. Each episode lasts for 3600 seconds at the end of which the actor and critic networks are updated with experiences collected from all the environments.
The learning curve demonstrates a clear and stable learning progression of the model. Initially, the model exhibits high variance and poor performance due to random exploration, reflected in large negative rewards. However, as training progresses, the agent consistently improves its policy, leading to a steady increase in average reward. The variability in rewards diminishes over time, reflecting increased policy stability and more consistent decision-making across environments. Full convergence is reached after around 2200 episodes. The smooth convergence of the learning curve toward near-zero reward further suggests that the agent reaches a stable and near-optimal policy, effectively minimizing traffic delay while maintaining robust performance throughout training.

\subsection{Comparative Performance Analysis: Impact of Attention Mechanism}
To evaluate operational efficiency and policy robustness, the proposed entity-centric RL model with hierarchical attention (RL-Att) was benchmarked against two baseline systems: an optimized Actuated Signal Control (ASC) plan configured via Synchro and a standard RL model utilizing a flat vector input state space without attention (RL-NoAtt). RL-NoAtt was formulated and trained on the same volume patterns as RL-Att in an earlier study \cite{53}. Performance was rigorously analyzed over N=10 independent microscopic simulation runs across two distinct traffic demand scenarios featuring structural volume variations completely unseen during model training.
\subsubsection{Scenario 1: Balanced Baseline Demand }
\label{sec:balanced_vols}
\begin{figure}[!h]
    \centering
    \includegraphics[width=1\linewidth]{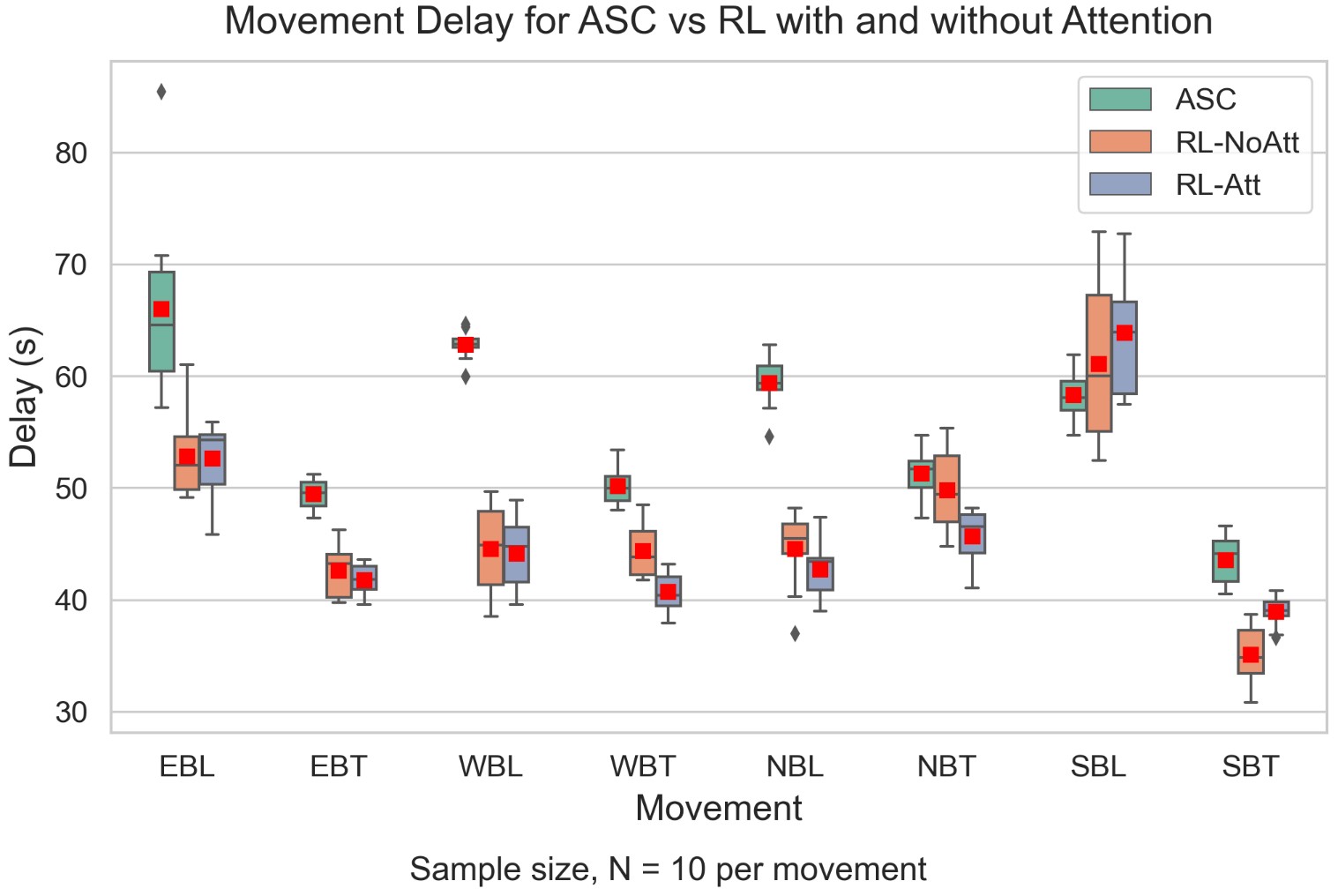}
    \caption{Comparing the performance of RL with and without Attention against optimized ASC}
    \label{fig:fig_5}
\end{figure}

The first validation is performed on a volume set with $V_{EBT}$=800, $V_{EBL}$=250, $V_{WBT}$=666, $V_{WBL}$=150, $V_{NBT}$=500, $V_{NBL}$=180,  $V_{SBT}$=400 and $V_{SBL}$=250 where $V_i$ is the volume in veh/h for movement i. As illustrated in Fig.~\ref{fig:fig_5}, both RL-based frameworks demonstrate substantial, statistically significant reductions in movement-level vehicular delay relative to the optimized ASC baseline across virtually all approaches, with the sole exception of SBL movement.

This systemic improvement is attributed to the contrast between the rule-based threshold logic of conventional ASC and the adaptive, state-dependent policies learned by the RL agents. The RL models optimize phase selections and green time distributions holistically by evaluating the concurrent traffic state profiles of all conflicting movements. Notably, the attention-infused framework (RL-Att) delivers comparable or superior delay reduction relative to RL-NoAtt, particularly for WBT, NBL and NBT. This indicates that the inclusion of an explanatory mechanism does not induce a performance penalty.
\subsubsection{Scenario 2: Highly Asymmetric Demand}
\begin{figure}[!h]
    \centering
    \includegraphics[width=1\linewidth]{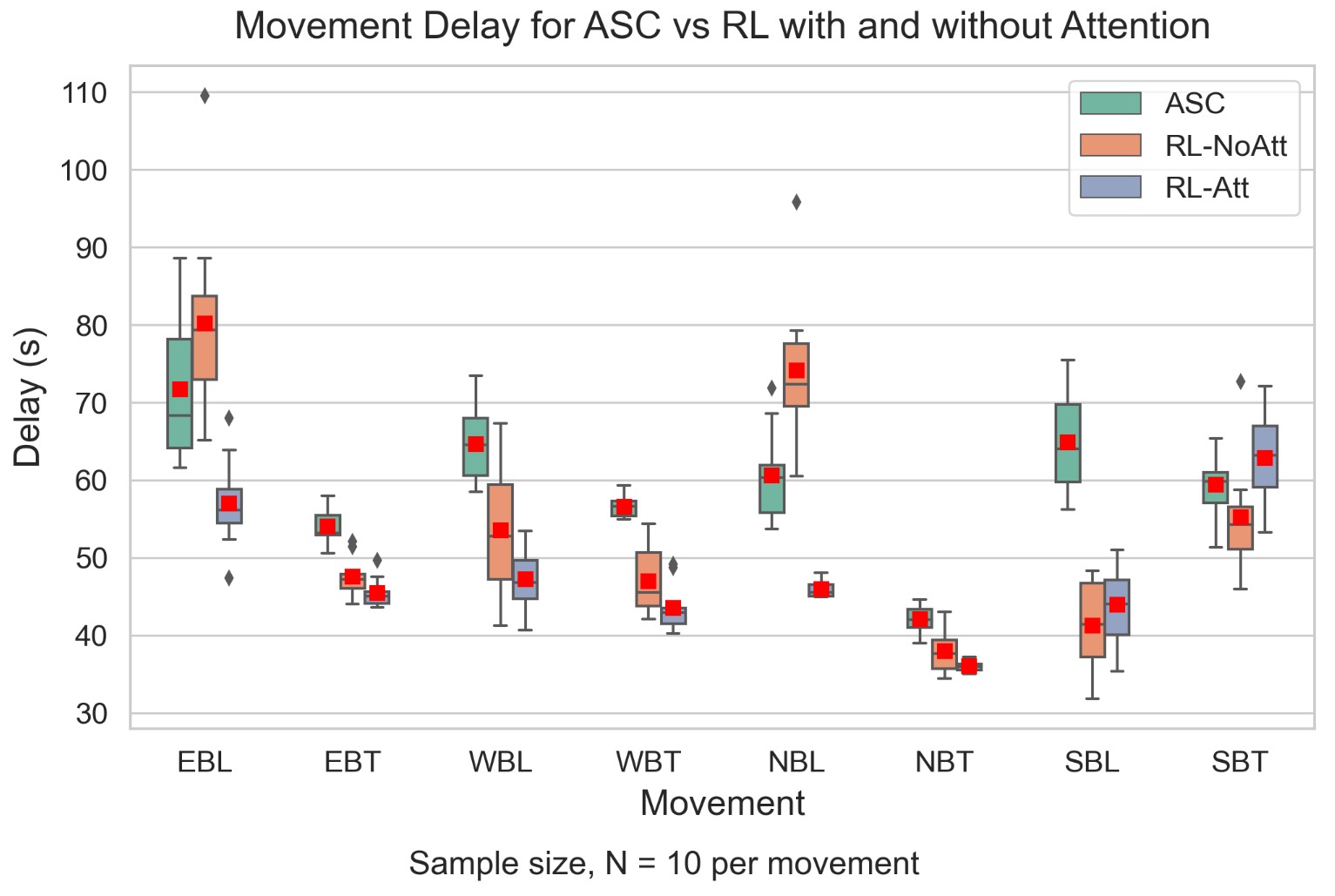}
    \caption{Comparing the performance of RL with and without Attention against optimized ASC on a highly imbalanced volume}
    \label{fig:fig_6}
\end{figure}

To further evaluate the structural robustness of the learned policies, a second scenario was introduced, specifically inflating the northbound left-turn demand (NBL increased from 180 to 432 veh/h). Southbound through (SBT) is increased to 400 veh/h while SBL is set to 30 veh/h. The empirical results for this highly asymmetrical O-D pattern are presented in Fig.~\ref{fig:fig_6}. Under this demand pattern, RL-Att consistently outperforms RL-NoAtt across all critical movements, showing minor degradation only on SBT movement. Most notably, for the highly congested NBL movement, RL-NoAtt exhibits an accumulation of delay, failing to efficiently adapt to the severe local demand pressure. Conversely, RL-Att dynamically restructures its internal phase sequencing and green-time allocation.
By utilizing entity-centric projections, RL-Att isolates specific lane demands and maps their relational dependencies to phase elements. The results suggest that this mathematical structure allows the agent to implicitly capture complex spatial-temporal interactions between geometric lane bottlenecks and active signal phases. Consequently, the attention-weighted architecture appears to anticipate local congestion spillback and proactively extends green intervals, demonstrating superior generalization, robustness, and stability in highly volatile deployment environments.
\subsection{Interpretability Analysis and Attention Field Dynamics}
To validate the transparent decision-making logic of the agent, the internal latent variations of the hierarchical attention blocks were captured and mapped directly against real-time traffic demand states.

% --- First Part: Bottom of Page 1 ---
\begin{figure*}[!b]
\centering
    \subfloat[Lane--phase attention weights for $t=2212$--$2258$.\label{fig_first_case}]{%
        \includegraphics[width=0.95\linewidth]{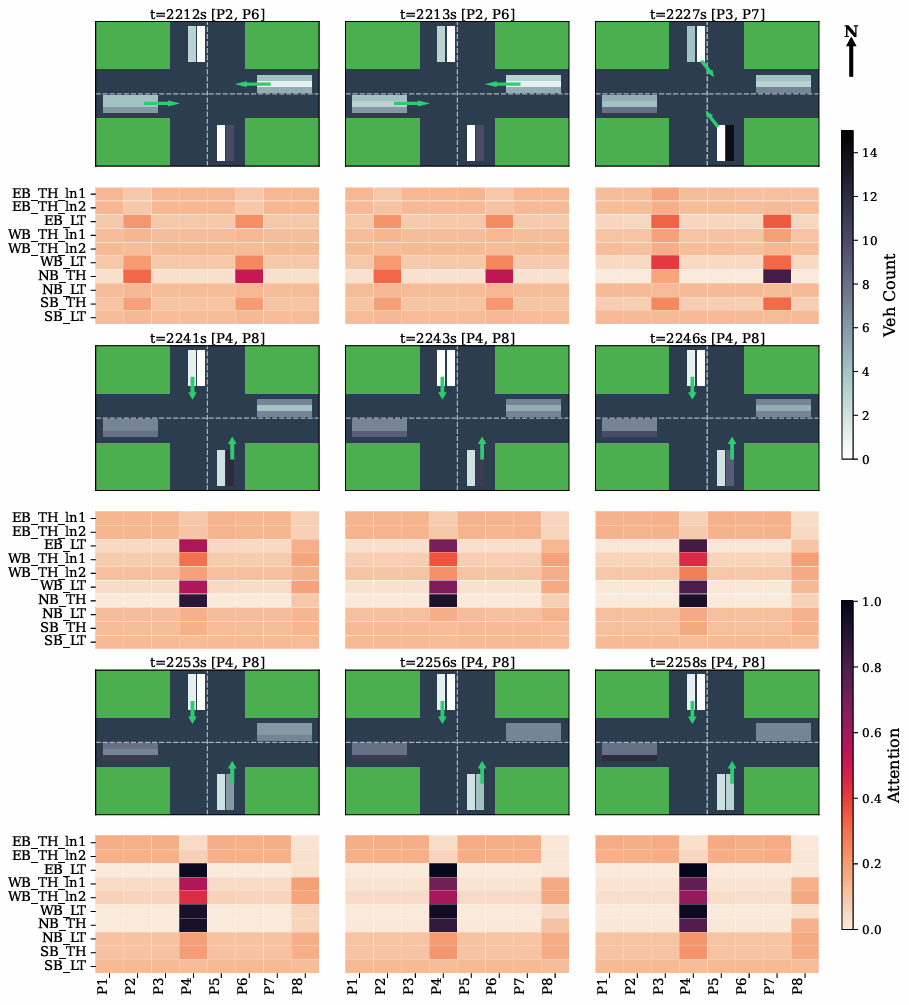}%
    }
    \phantomcaption % <--- Tells LaTeX to advance the counter to Fig. 7 right here!
\end{figure*}

% --- Second Part: Top of Page 2 ---
\begin{figure*}[!t]
\centering
    \ContinuedFloat % Now successfully holds it on Fig. 7 and sets this subfloat to (b)
    \subfloat[Lane--phase attention weights for $t=2271$--$2310$.\label{fig_second_case}]{%
        \includegraphics[width=0.95\linewidth]{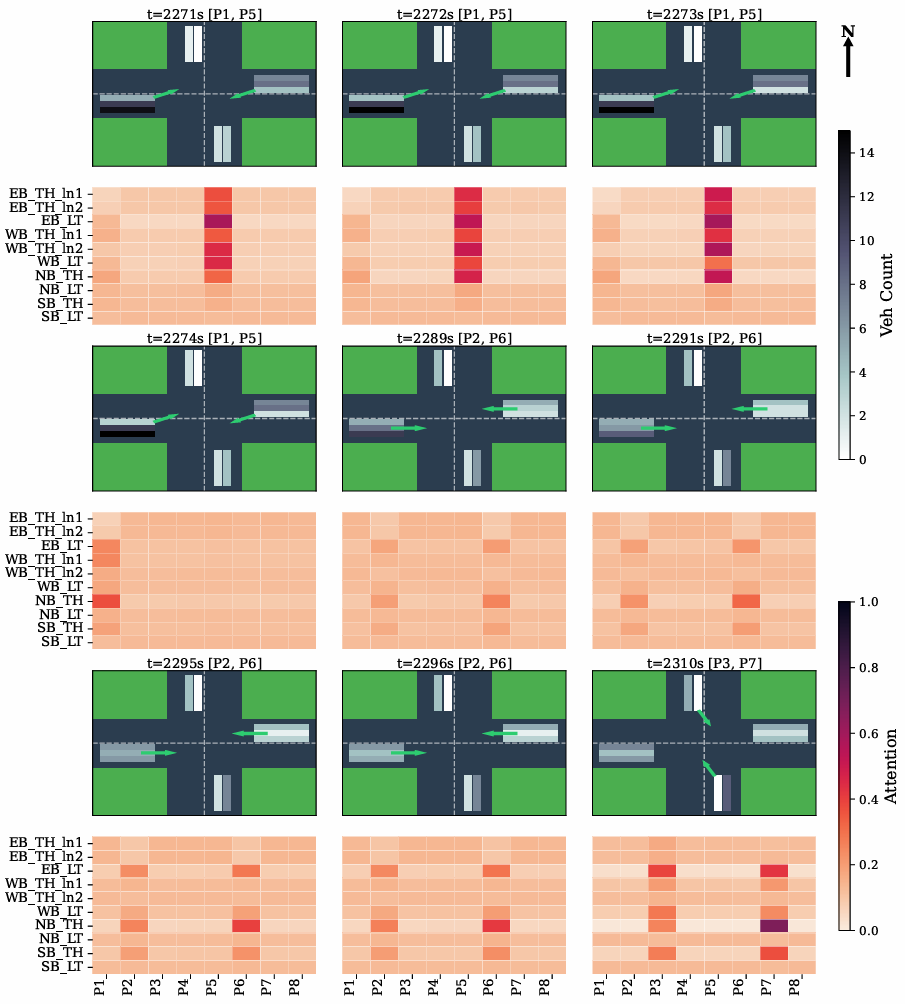}%
    }
    \caption{Lane--phase cross-attention weights and corresponding active vehicle counts. (a) Weights for interval $t=2212$--$2258$. (b) Weights for interval $t=2271$--$2310$.}
    \label{fig_7}
\end{figure*}
\subsubsection{Lane--Phase Cross Attention Dynamics}
\label{sec:lane_phase}
Fig.~\ref{fig_7} illustrates the temporal evolution of lane--phase attention alongside vehicle counts in each lane over a representative simulation interval ($t = 2221$--$2310$\,s). The test volumes are the same as in Section~\ref{sec:balanced_vols} above. Active phases are denoted by green arrows and explicit phase markers in square brackets. Because the control framework operates as an event-driven SMDP, decision intervals are non-uniform and occur dynamically based on current signal states and selected actions. Consequently, attention weights are logged exclusively at active policy execution points; intermediate phase preservation loops (such as the sustained execution of \{P4, P8\} between $t = 2241$ and $t = 2258$\,s) are omitted for analytical clarity.

An empirical trace of the sequence reveals a highly intuitive alignment with domain-specific traffic control logic:
\begin{itemize}
    \item At the beginning of the sequence ($t = 2221$\,s), phases \{P2, P6\} are active, serving the east--west through movements. As a queue builds up on the northbound through lane (\texttt{NB\_TH}), its query projection assigns a progressively elevated attention weight to the active phases, demanding service.
    \item At $t = 2227$\,s, the model initiates a phase transition, from \{P2, P6\} first passing through \{P3, P7\} and subsequently activating phases \{P4, P8\}, which serve the north--south movements. The attention associated with \texttt{NB\_TH} remains elevated during this period, consistent with the presence of sustained queuing, and gradually decreases as the queues dissipate.
    \item As the \texttt{NB\_TH} queue dissipates, demand rebuilds along the east--west approaches, particularly for the through and left-turn movements (\texttt{EB\_TH}, \texttt{WB\_TH}, \texttt{EB\_LT}, and \texttt{WB\_LT}). The cross-attention matrix tracks this shift, raising the weights for these lanes on the active phases.
    \item This pattern repeats over the remainder of the interval, where increases in lane-level demand are followed by elevated attention weights and subsequent activation of the associated phases. Across the sequence, phase transitions consistently occur when attention becomes concentrated on alternative phase groups, indicating a strong temporal alignment between demand buildup, attention allocation, and control actions.
\end{itemize}
% --- First Part: Bottom of Page 1 ---
\begin{figure*}[!b]
\centering
    \subfloat[Lane--Lane attention weights for $t=603$--$646$.\label{fig_first_case_8}]{%
        \includegraphics[width=0.95\linewidth]{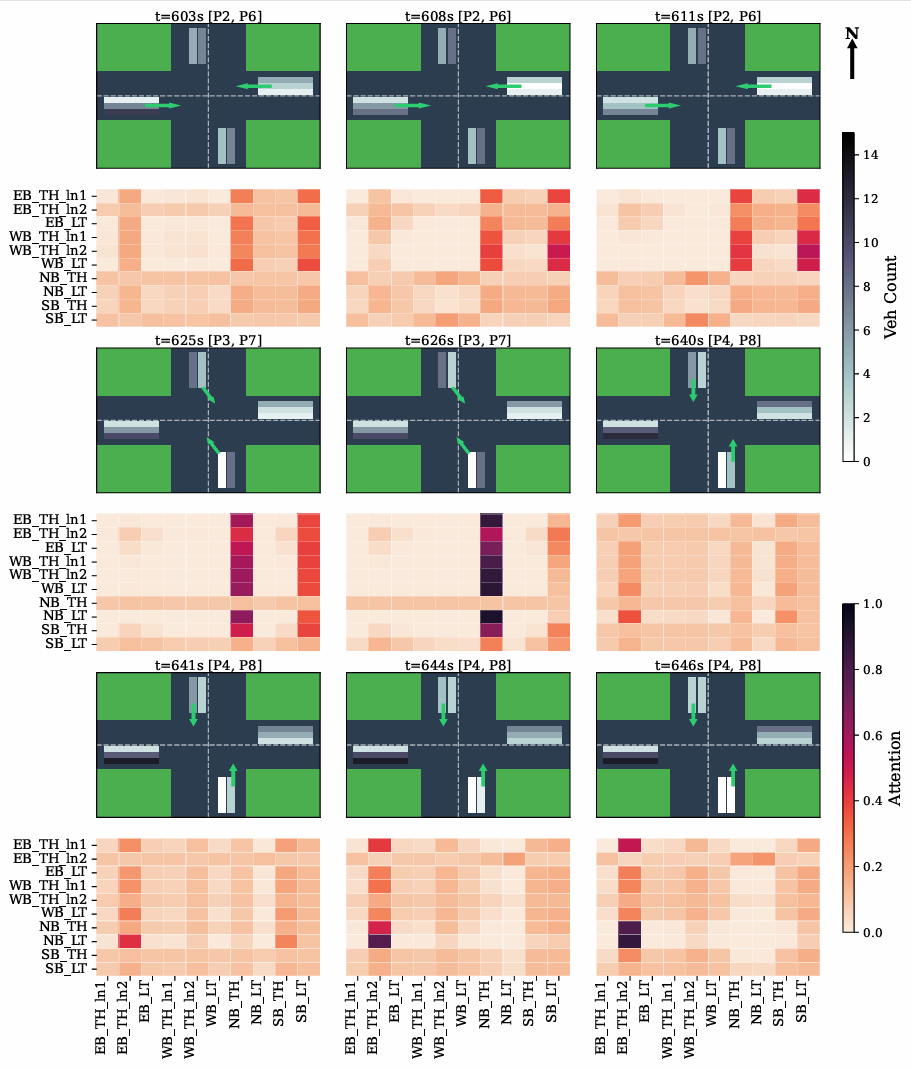}%
    }
    \phantomcaption % <--- Increments the counter to Fig. 8 correctly without printing text
\end{figure*}

% --- Second Part: Top of Page 2 ---
\begin{figure*}[!t]
\centering
    \ContinuedFloat % Holds the counter on Fig. 8 and sets this subfloat to (b)
    \subfloat[Lane--Lane attention weights for $t=659$--$707$.\label{fig_second_case_8}]{%
        \includegraphics[width=0.95\linewidth]{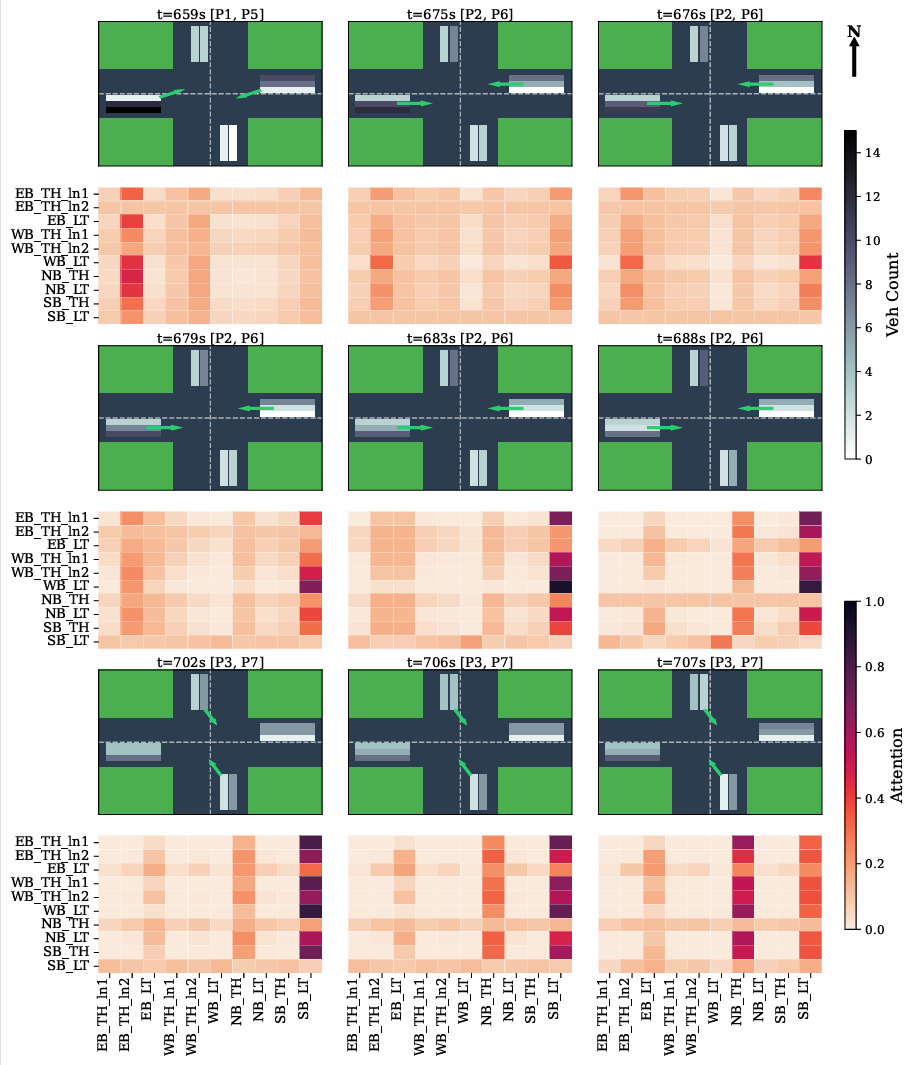}%
    }
    % Final unified master caption
    \caption{Lane--lane self-attention weights and corresponding active vehicle counts under balanced traffic demand. (a) Weights for $t=603$--$646$. (b) Weights for $t=659$--$707$.}
    \label{fig_8}
\end{figure*}
This tight temporal alignment between spatial queue buildup, attention weight concentration, and phase triggering demonstrates that the cross-attention mechanism yields a valid, human-interpretable explanation of policy inference.

\subsubsection{Lane--Lane Self-Attention Under Balanced Traffic Demand}
Fig.~\ref{fig_8} presents the lane--lane attention weights over a representative simulation interval. In this formulation, each lane attends to other lanes, allowing the model to capture spatial interactions and dependencies across the intersection. The test demand scenario consists of balanced traffic conditions. Similar to the lane--phase analysis in Section~\ref{sec:lane_phase} above, only representative decision points are shown, with non-critical intermediate steps omitted for brevity.
\begin{figure*}[!t]
    \centering
    \includegraphics[width=\linewidth]{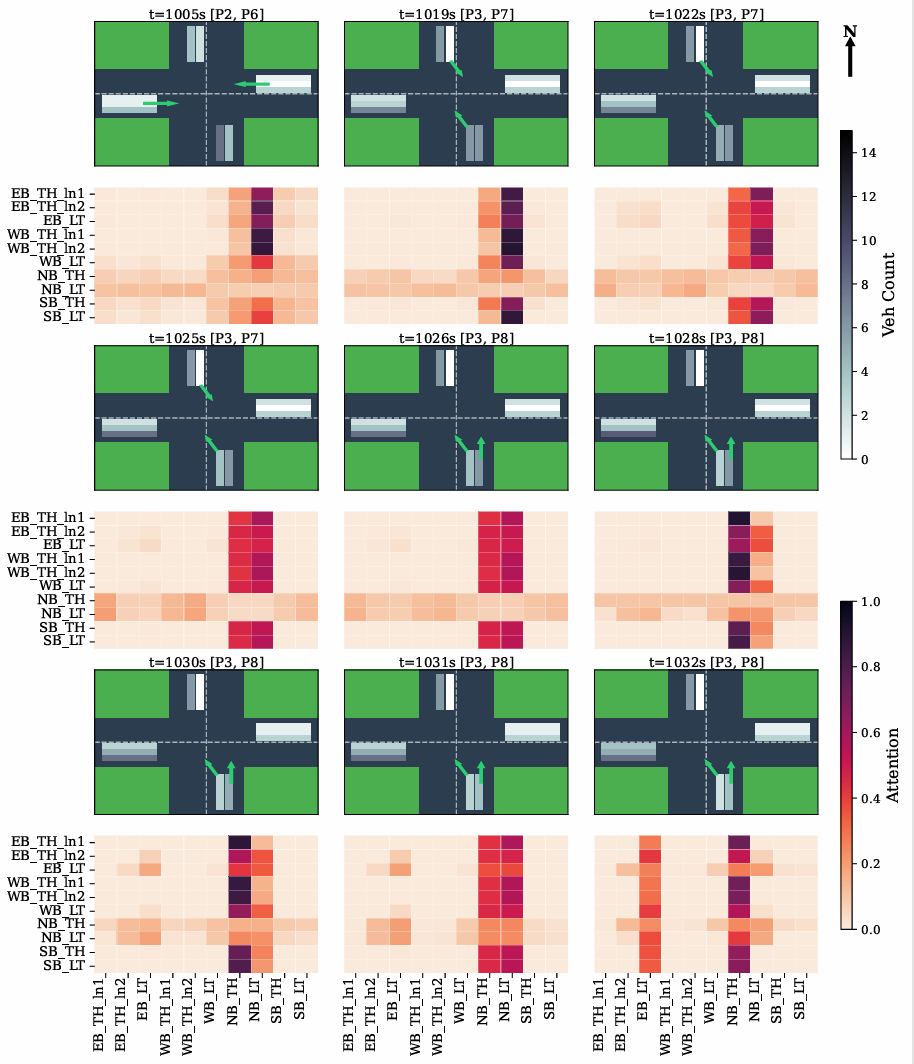}
    \caption{Lane--lane self-attention weights under asymmetric traffic demand.}
    \label{fig:fig_9}
\end{figure*}
A temporal examination of Fig.~\ref{fig_8} reveals how lane--lane attention evolves in response to changing demand patterns and phase transitions. At the start of the interval ($t = 603$\,s), phases \{P2, P6\} are active, serving main \texttt{EB} and \texttt{WB} through movements. Queues are building up on side street lanes, and \texttt{NB\_TH} and \texttt{SB\_LT} lanes receive the highest attention from the opposing movement lanes. The self-attention mechanism establishes \texttt{NB\_TH} and \texttt{SB\_LT} as central informational anchors across the entire layer, forcing currently served movements to continuously evaluate their remaining green time against the mounting vehicle pressure of these specific conflicting flows. This look-ahead capability directly registers the urgent need for a service switch, guiding the policy's capacity to handle phase transitions smoothly.

Consequently, at $t = 611$\,s, a phase transition occurs to phases \{P3, P7\}, and subsequently to phases \{P4, P8\} at $t = 626$\,s. While phases \{P4, P8\} are active, queues start to build up again on the main street, and correspondingly, attention starts to shift back to the main street lanes tracking the emerging demand pressure. A similar pattern is observed throughout the remainder of the sequence, where attention intensifies on lanes experiencing increased demand and subsequently redistributes following phase changes. 

Throughout the interval, attention becomes more concentrated on specific lanes during periods of high demand and more diffuse when traffic conditions are balanced, closely tracking the temporal evolution of traffic states and control actions. The attention patterns reveal a clear structure in which lanes with higher vehicle counts exert stronger influence on other lanes, particularly those served by a conflicting phase. Notably, the attention weights concentrate on a subset of critical lanes, suggesting that the model selectively focuses on the most influential traffic streams at each decision point. These patterns evolve over time in response to changing demand, demonstrating that the model learns dynamic, context-dependent relationships between lanes rather than static correlations.

\subsubsection{Lane--Lane Attention Self-Attention under asymmetric traffic Demand}
To further examine the internal representations learned by the model, a highly asymmetrical scenario is introduced by inflating the northbound approach volumes ($V_{\text{NBT}} = 500$, $V_{\text{NBL}} = 432$) while limiting the opposing southbound left-turn volume ($V_{\text{SBL}} = 30$). The tracking interval between $t = 960$\,s and $t = 1032$\,s is shown in Fig.~\ref{fig:fig_9}.

Despite this imbalance in volumes, the trained model demonstrates strong robustness by effectively allocating green time to the northbound movements and preventing excessive queue buildup. For the \texttt{NB} approach, phases \{P3, P8\} are frequently paired during periods of sustained demand (only $t = 1026$\,s to $t = 1032$\,s shown here). This behavior is particularly important given the limited storage capacity of the northbound left-turn lane. By consistently serving phase P3 (\texttt{NB} left-turn) in coordination with P8 (\texttt{NB} through), the controller limits queue growth and prevents spillback that could block the turn bay or interfere with through traffic.

This coordinated/joint phase selection is reflected in the lane--lane attention patterns, where the \texttt{NB} left-turn lane maintains elevated attention not only within its own movement but also in relation to the \texttt{NB} through lane. This suggests that the model has implicitly learned to account for geometric constraints and potential spillback effects, using coordinated phase activation to mitigate the risk of lane blockage. More broadly, the attention patterns exhibit a sustained focus on the northbound lanes throughout the interval. Unlike the more dynamic attention shifts observed under balanced conditions, the attention weights remain consistently concentrated on these movements, reflecting persistent demand pressure. This behavior indicates that the model adapts its internal representation to prioritize dominant traffic streams while maintaining stable and efficient control without oscillatory phase switching.

\section{Conclusions}
This effort formulates an explainable, entity-centric RL framework designed to overcome the critical trust and transparency barriers that have that have the potential to restrict the real-world deployment of deep RL in traffic signal control. By abandoning flat state vectors in favor of an architecture that explicitly treats intersection components as independent high-dimensional lane and phase entities, the proposed model successfully opens the "black box" of deep RL neural networks.

The hierarchical attention mechanism, combining multi-head cross-attention and self-attention delivers a real-time explainability output affinity matrix. This enables operators to visually and analytically audit how the network weights competing lane demands against phase configurations. Furthermore, the integration of a deterministic action-masking interface successfully bridges the gap between stochastic policy exploration and critical field safety mandates, ensuring absolute compliance with NEMA ring-barrier, minimum green-time and clearance interval constraints.

Empirical evaluations conducted within the SUMO microscopic traffic simulation environment validate that this injection of transparency does not come at the expense of operational performance. The framework matches or outperforms state-of-the-practice baselines in delay minimization while inherently mitigating structural traffic anomalies such queue spillbacks, turn lane blockages and asymmetric congestion bottlenecks. More importantly, the corresponding attention maps demonstrate intuitive alignment with established traffic engineering principles. By providing human-interpretable diagnostics, this architecture lays the necessary foundation for agency acceptance, operational trust, and collaborative fine-tuning by domain experts, moving adaptive traffic signal control a significant step closer to widespread deployment.

\bibliographystyle{IEEEtran}

\section*{Biography Section}

\begin{IEEEbiographynophoto}{Dickens Kwesiga}
is a Research Engineer with the School of Civil and Environmental Engineering, Georgia Institute of Technology, Atlanta, GA, USA. His current research focuses on developing deployment-oriented artificial intelligence (AI) systems for Intelligent Transportation Systems (ITS), validated through software-in-the-loop (SIL) and hardware-in-the-loop (HIL) simulation testing to ensure real-world readiness.
\end{IEEEbiographynophoto}

\begin{IEEEbiographynophoto}{Nishu Choudhary}
is a Research Engineer with the School of Civil and Environmental Engineering at the Georgia Institute of Technology. Her research interests include leveraging machine learning (ML) techniques to develop data-driven solutions for proactive traffic management and intelligent transportation systems.
\end{IEEEbiographynophoto}

\begin{IEEEbiographynophoto}{Angshuman Guin}
is a Senior Research Engineer with the School of Civil and Environmental Engineering, Georgia Institute of Technology, Atlanta, GA, USA. His research interests include freeway operations, connected and autonomous vehicles, Intelligent Transportation Systems (ITS), transportation safety, traffic simulation, and transportation data management.
\end{IEEEbiographynophoto}

\begin{IEEEbiographynophoto}{Michael Hunter}
is a Professor with the School of Civil and Environmental Engineering, Georgia Institute of Technology, Atlanta, GA, USA. His teaching and research interests include transportation operations and design, with emphasis on emerging transportation technologies, adaptive traffic signal control, traffic simulation, and arterial corridor operations.
\end{IEEEbiographynophoto}

%\vfill

\end{document}